\documentclass[openacc]{rstransa}

\newcommand{\alg}{SymRegg}

\usepackage{subfig}

\titlehead{Research}

\begin{document}

\title{Equality Graph Assisted Symbolic Regression}

\author{
Fabricio Olivetti de Franca$^{1}$ and Gabriel Kronberger$^{2}$}

\address{$^{1}$Federal University of ABC, Santo Andre, SP, Brazil\\
$^{2}$University of Applied Sciences Upper Austria, Hagenberg, Austria}

\subject{xxxxx, xxxxx, xxxx}

\keywords{symbolic regression, equality saturation, equality graphs}

\corres{Fabricio Olivetti de Franca\\
\email{folivetti@ufabc.edu.br}}

\begin{abstract}
In Symbolic Regression (SR), Genetic Programming (GP) is a popular search algorithm that delivers state-of-the-art results in term of accuracy.
Its success relies on the concept of neutrality, which induces large plateaus that the search can safely navigate to more promising regions. 
Navigating these plateaus, while necessary, requires the computation of redundant expressions, up to $60\%$ of the total number of evaluation, as noted in a recent study. 
The equality graph (e-graph) structure can compactly store and group equivalent expressions enabling us to verify if a given expression and their variations were already visited by the search, thus enabling us to avoid unnecessary computation.
We propose a new search algorithm for symbolic regression called SymRegg that revolves around the e-graph structure following simple steps: perturb solutions sampled from a selection of expressions stored in the e-graph, if it generates an unvisited expression, insert it into the e-graph and generates its equivalent forms.
We show that SymRegg is capable of improving the efficiency of the search, maintaining consistently accurate results across different datasets while requiring a choice of a minimalist set of hyperparameters.
\end{abstract}


\begin{fmtext}
\section{Introduction}

Discovering mathematical equations from observations is an important step for understanding scientific laws. Symbolic regression (SR)~\cite{Koza1992,kronberger2024} automates this step by searching the space of mathematical functions and returning a set of alternative hypotheses for the studied law.
The current literature~\cite{kronberger2024,udrescu2020ai,russeil2024multiview} already presents different success cases of applying SR in either discovering new governing laws or finding better alternatives for existing ones.

Genetic programming (GP) is currently the main
\end{fmtext}
\maketitle
\noindent 
search heuristic among the top performing SR algorithms~\cite{la2021contemporary,de2024srbench} characterized as a \textit{population-based} approach that searches multiple hypotheses in parallel while either perturbing or combining good solutions.
The general idea is that the selective pressure will favor the maintenance of well performing building blocks in the population with the expectation that random variations or recombination of those will reflect in the quality of the newly generated solution.
One of the issues of GP is that it relies on the selective pressure to guide the process to the global optima, which is reasonable under two circumstances: i) the solution representation has little or no \textbf{redundancy}, and ii) a small change in the solution mostly reflects in a small change in the quality of the perturbed solution, making the \textbf{navigation} through the search space predictable. Unfortunately, this is not true for SR as any expression can have infinitely many equivalent expressions, and a simple change of one symbol of the expression (i.e., replace a $\log$ with $\exp$), will reflect into some major change in the behavior of the function. 

This redundancy was studied  by Kronberger et al.~\cite{kronberger2024inefficiency} showing that, during the search, up to $60\%$ of the expressions visited by GP are, in fact, equivalent to previously visited ones. This has consequences when the function contains numerical parameters that must be fitted to the data reducing the probability of finding their optimal values~\cite{de2023reducing,kronberger2024jsc}.
This redundancy can actually be beneficial to alleviate the navigability problem, as pointed out in~\cite{keller1999evolution,milleratall2006} as it increases the size of the set of neighbors of a solution,  thus increasing the probability of reaching an improved equation after some perturbation steps.
Nevertheless, visiting redundant expressions wastes computational resource that could be used to evaluate distinct alternatives. 

To avoid redundant evaluations, it is possible to derive all equivalent forms of a certain expression using a set of equivalence rules with the algorithm called equality saturation~\cite{willsey2021egg}. This algorithm uses the equality graph (e-graph) data structure to compactly represent the equivalence relationship of expressions.
This technique can be combined in the equation search procedure to allow the perturbation of all the equivalent expressions without the need to evaluate them. 

In this paper, we introduce \alg{}, a heuristic search that exploits the e-graph and equality saturation to iteratively build a database of visited expressions and their equivalent forms while exploiting this knowledge base to guide the search towards good quality hypotheses that accurately describe the observed data.
As a highlight, \alg{} is not a population-based approach and requires the adjustment of only a few hyperparameters, unlike current GP implementations that often supports a large number of hyperparameters that must be correctly set in order to achieve the best performance. We show in the results section that, despite the simple-to-use experience, \alg{} explores the search space more efficiently than GP. 

This paper is organized as follows: in section~\ref{sec:eqdiscovery} we briefly explain the basic concepts of equation discovery, section~\ref{sec:eqsat} introduces the general idea of equality graphs and equality saturation. In section\ref{sec:symregg} we explain the \alg{} algorithm and how we can exploit the information store in the equality graphs. Sections~\ref{sec:experiments}~and~\ref{sec:results} describes the experimentation methods and obtained results showing the efficiency of \alg{}. We conclude the paper with some final dicussions in section~\ref{sec:conclusion}.

\section{Equation Discovery}\label{sec:eqdiscovery}

The equation discovery problem can be formulated as the search for a function $f(x; \theta)$ that approximates observations $\{x^{(i)}, y^{(i)}\}_{i=1 \ldots n}$ obtained through careful experiments. In this formulation, $x^{(i)} \in \mathbb{R}^d$ is often a $d$-dimensional real-valued vector where each $x^{(i)}_j$ is called a predictor or feature, $\theta \in \mathbb{R}^k$ is a $k$-dimensional vector representing adjustable parameters adding degree-of-freedom to the function that improves the goodness-of-fit and can make it a more general description of the studied phenomena. These parameters must be optimized using a loss function dependent of the available data.
Although in this description we are restricting the predictors to real-valued vectors, it is also possible to have features on different domains by either performing transformations or adapting the search algorithm.

Beyond this data-only formulation, it is also possible to include additional desiderata in the form of expert knowledge or expectation about the behavior of the function, such as shape-constraints~\cite{kronberger2022shape} \textcolor{red}{and unit information~\cite{keijzer1999dimensionally} (or help performing unit inference of adjustable parameters~\cite{reuter2024unit}).}
There are many different search methods available for this purpose with different characteristics such as being deterministic~\cite{mcconaghy2011ffx, de2018greedy}, exhaustive~\cite{bartlett2023exhaustive,kammerer2020symbolic}, using neural network~\cite{de2021interaction, landajuela2022unified}, but the current state-of-the-art algorithms rely on GP~\cite{la2021contemporary,de2024srbench} as the search method \textcolor{red}{when considering error minimization or the error-complexity tradeoff}.

GP belongs to the family of meta-heuristics inspired by the natural evolutionary process. The general idea is to start the search process with a \emph{population} of randomly generated candidate solutions and iteratively perturbing and combining them until it reaches a certain stop criteria. The perturbations are performed on a selected subset of the current population where the probability of being selected is proportional to the quality of the candidate solution, this adds a bias towards the exploration of good regions of the search space. The combination of two solutions follows from the expectation that good solutions have a higher probability of containing good quality building blocks that can help generating the optimal equation.

The building blocks hypothesis~\cite{holland2000building} is closely connected to the optimal substructure property~\cite{held1962dynamic,bellman1962dynamic} that relies on the optimality of separate components of a problem to correspond to the optimal solution of the full problem. Specifically for symbolic regression, this may not be the case as the behavior of a piece of a function depends on the context it is inserted into~\cite{o1995troubling}, thus even the apparently small perturbation can induce a big change in the function behavior.

Despite not having a stronger support for the building block hypothesis, equation discovery has an important property that allows the exploration of the search space: redundancy in representation. Any mathematical function can be represented in infinitely many forms by either applying equivalence relations (e.g., $x_1 + x_1 = 2 x_1$) or adding a neutral expression (e.g., $2x_1 + (x_2 / x_2 - 1) = 2x_1$). This redundancy, also referred to as neutrality~\cite{hu2018neutrality,banzhaf2024combinatorics}, allows the solution candidates to accumulate inactive building blocks throughout the search that can become active and relevant at a future step. This, however, is also a source of inefficiency of the search that, as pointed out by Kronberger et al.~\cite{kronberger2024inefficiency}, can lead to the evaluation of up to $60\%$ of previously visited expressions in SR search spaces with expression lengths limited to less than 15 symbols. Under the perspective of the exploration-exploitation tradeoff, neutrality promotes the exploration of the search space, allowing the algorithm to incrementally increase the neighborhood region with the redundant building blocks. On the other hand, the lack of a mechanism for controlling the introduction of such redundancies creates plateaus in the search space, limiting its capability for exploitation.  
The main challenge is how to detect equivalent expressions efficiently so that the algorithm can not only detect duplicates but also ensure the production of new expressions. This can be accomplished by the use of equality graphs as explained next.

\section{Equality Graphs}\label{sec:eqsat}

\begin{figure}[t!]
    \centering
    \subfloat[]{\includegraphics[width=0.3\linewidth]{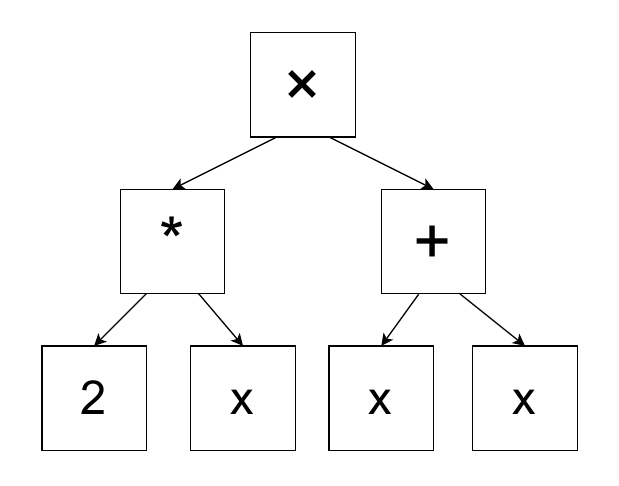}\label{fig:tree}} 
    \subfloat[]{\includegraphics[width=0.21\linewidth]{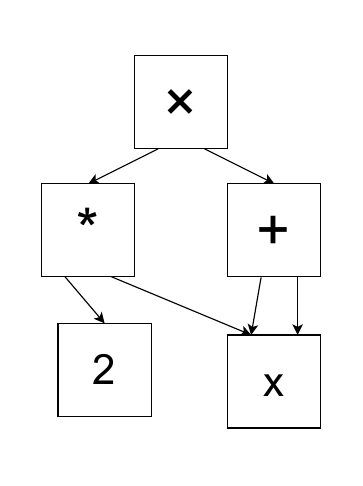}\label{fig:dag}}
    \subfloat[]{\includegraphics[width=0.18\linewidth]{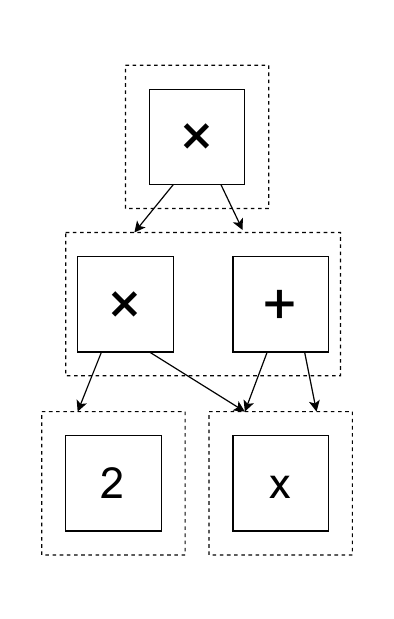}\label{fig:egraph}}
    \caption{The expression $2x(x+x)$ represented as (a) a tree, (b) a directed acyclic graph and, (c) an e-graph. The e-graph adds dashed lines around the nodes that correspond to equivalent expressions.}
\end{figure}

Any mathematical expression can be represented as a tree data structure (Fig.~\ref{fig:tree}) in which each node represents a symbol, and the non-terminals nodes point to other nodes. This data structure explicitly keeps the repeated branches (e.g., multiple occurrences of $x$) and evaluate them independently. A more compact form can be obtained when using a directed acyclic graph (dag, Fig.~\ref{fig:dag}) which reuses shared components avoiding recalculating repeated parts of the expression, but still cannot assert equivalence relations (e.g., $2x$ and $x+x$).  
Equality graphs (e-graphs)~\cite{tate2009equality} extend the concept of a graph by allowing the grouping of nodes (hereby called e-nodes) expressing equivalent relations (Fig.~\ref{fig:egraph}). In this context, the e-graph is composed of a set of \emph{e-nodes} that can represent a terminal or a non-terminal symbol, a set of \emph{e-classes} (dashed lines in the example) that groups together equivalent e-nodes, and a set of \emph{edges} connecting non-terminal e-nodes to e-classes. In the illustrative example, both $2x$ and $x+x$ are grouped together as they are equivalent and are evaluated to the same values for every value of $x$. One immediate advantage of this representation is that, whenever an expression contained in a given e-class is evaluated, it is not necessary to evaluate all the equivalent expressions, thus saving computation.

Whenever the e-graph is non-empty, it is possible to apply the algorithm called \emph{equality saturation}~\cite{willsey2021egg} to generate and group all equivalent relations.
Conceptually, the algorithm follows these simple steps: i) match all the equivalence rules in the current state of the e-graph, ii) apply the rules creating new e-classes and e-nodes, iii) merge the equivalent e-classes, iv) repeat until saturation.
The main bottlenecks of this algorithm is the pattern matching of equivalence rules, which requires the maintenance of a database of patterns for fast retrieval, and the potential exponential growth of the e-graph, since some patterns can be expanded indefinitely.

Besides the node labels, the e-classes can hold additional information that can be deduced from the e-graph or using external knowledge. For example, each e-class can contain the size of the smaller expression among all stored equivalences, the partial evaluation given a dataset, whether that expression is always positive, and the physical unit information. These properties are deduced as the e-graph is built by propagating the information upwards\footnote{Assuming that the root of the tree is at the top, something every computer scientist believes to be.}. For example, if you state that the unit for $x$ is $m$, when we insert $x^2$ into the e-graph, this e-class will contain the unit $m^2$, additionally, we can assert that this e-class will always evaluate to positive values and it will be a concave up function. 
Considering the combination of the e-graph and the pattern database as a knowledge graph built incrementally during the search procedure, we can exploit the stored information to improve the search.

\begin{figure}[t!]
    \centering
    \subfloat[]{\includegraphics[width=0.39\linewidth]{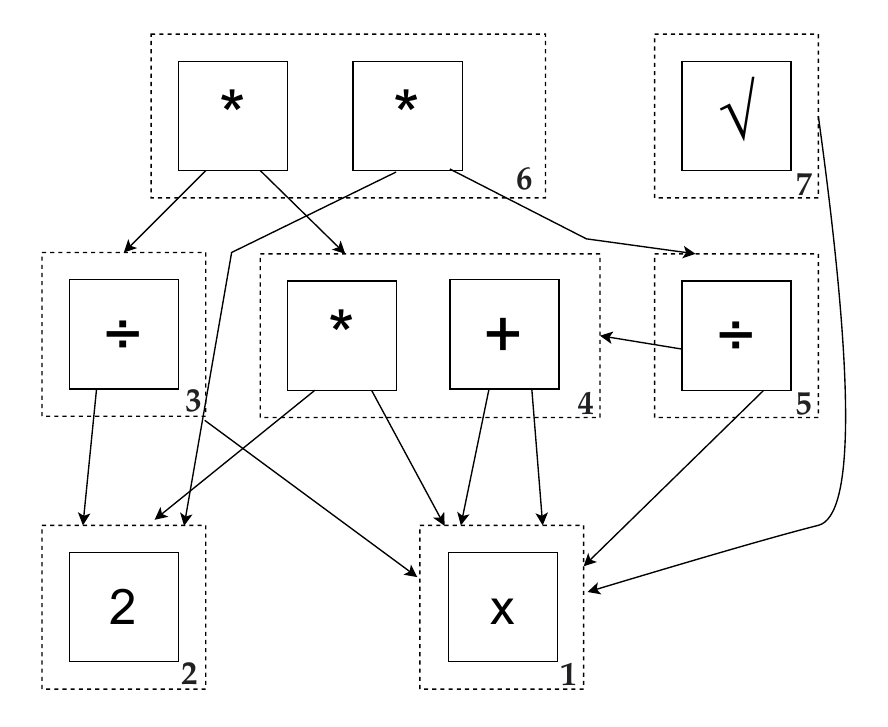}\label{fig:egraph2a}} 
    \subfloat[]{\includegraphics[width=0.45\linewidth]{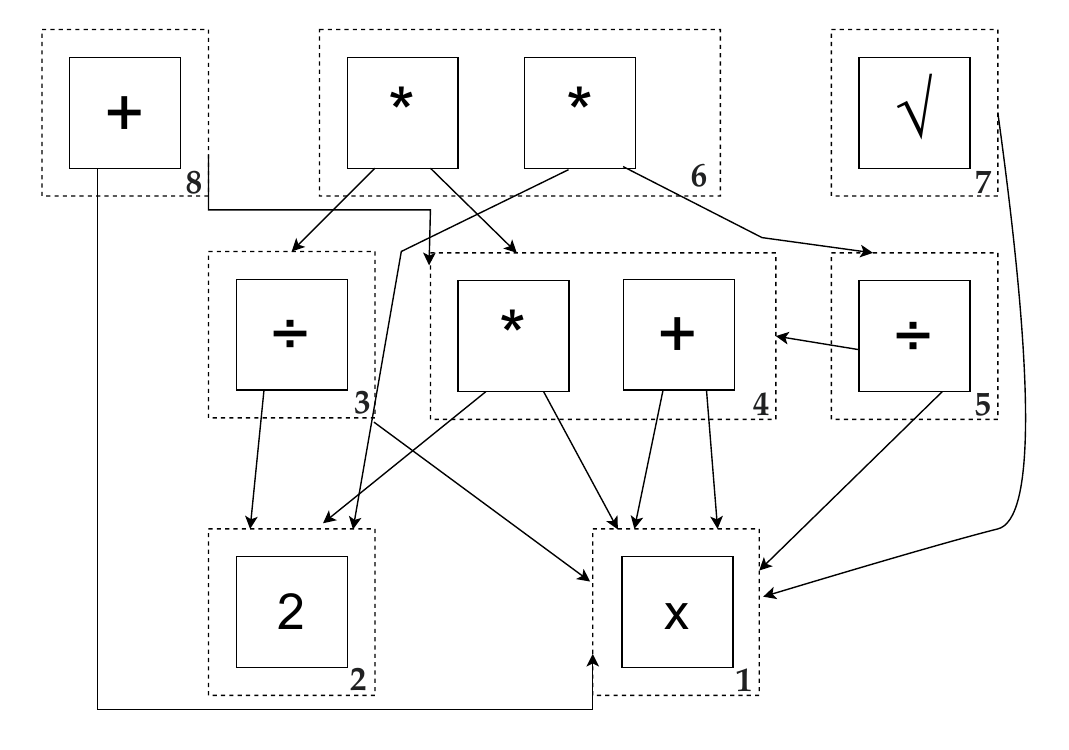}\label{fig:egraph2b}}
    \caption{(a) Illustrative example of an e-graph. Solid boxes represent e-nodes and dashed lines represent e-classes (id numbers in the lower right). Extracting expressions by following any path of a given e-class will represent equivalent expressions. For example, expressions $2x$ or $x+x$, extracted from e-class $4$. When inserting the expression $x + 2x$ (b), the already present e-classes will be reused (see e-class $8$).}
    \label{fig:egraph2}
\end{figure}

The common usage of this structure is to create an e-graph out of a single expression, applying equality saturation to generate all the equivalent expressions, and then extracting the optimal expression using a cost heuristic. But, the definition of the structure enables the storage of multiple expressions assuming the existence of multiple roots, as long as the reference of which e-class represents a root is stored for bookkeeping. Fig.~\ref{fig:egraph2a} shows one example of creating a single e-graph from the expressions $(2/x)(x+x), 2x, \sqrt{x}$, in this case the root e-classes are $\{6,4,7\}$, represented by their corresponding ids (small number in the bottom right of each dashed box).

New expressions are added starting from the terminal nodes when the algorithm verifies whether they already exist in the e-graph. If they do, the e-class id is returned, otherwise a new e-class and e-node is created and the newly created id is returned.
After the terminals are inserted, the parents (non-terminals) are created by joining the token with the e-class ids of the children. In our example (Fig.~\ref{fig:egraph2}), the expression $2x$, would first retrieve the ids $2$ and $1$, corresponding to the two terminals, and create the expression $\underline{2} \times \underline{1}$\footnote{The underlined numbers represent the e-class ids.}. This is matched against a database of existing e-nodes, which will identify that it belongs to e-class $4$. Likewise, if the expression $x+x$ is inserted into the e-graph, the same e-class id ($4$) would be returned, expressing the equivalence of both expressions. This procedure continues until it reaches the root of the expression and nothing else needs to be inserted.
We can see from this description, that it is straightforward to detect whether a given expression exists in the e-graph or any of its equivalence form.
Of course the detection of equivalent expressions is only possible if all equivalences are generated during equality saturation, which is not often the case. But, applying a few iterations of equality saturation for each expression already extends the e-graph with a number of equivalent expressions.

Another feature of expression insertion in e-graph is constant folding, in which e-classes that evaluate to a constant are automatically pruned into a single e-node. For example, upon inserting the expression $2 + 3$, the constant $5$ would be inserted into the e-graph instead of the whole expression. The same idea can be applied to adjustable parameters, so an expression $\theta_1 x / \theta_2$ would be automatically replaced by $\theta_1 x$, as the second parameter is redundant for the data fitting process, this reduces the effects of  overparametrization~\cite{de2023reducing,kronberger2024jsc} in regression models, that can reduce the probability of success of the optimization method.

These two features of detecting whether an expression exists in the e-graph and avoiding the creation of overparameterized expressions are crucial to the development of search algorithms for equation discovery~\cite{de2023reducing,kronberger2024jsc,kronberger2024inefficiency}. In particular, Kronberger et al.~\cite{kronberger2024inefficiency} reported that genetic programming has a tendency of revisiting expressions or their equivalent form during the search. This is caused by the selective pressure that stimulates the convergence of the population of solutions to a selection of expressions with similar building blocks. During perturbation and recombination, this can lead to the creation of repeated expressions. Ting et al.~\cite{hu2018neutrality} and Wolfgang et al.~\cite{banzhaf2024combinatorics} argue that this is beneficial to the equation discovery process since it enables the application of multiple perturbation steps without a change in the goodness-of-fit. This, in turn, creates multiple \emph{exit points} from one expression to different regions of the search space, improving the exploration capabilities of the search method without the cost of evaluating redundant expressions. In the next section, we will propose a new search algorithm for equation discovery capable of exploiting the e-graph structure to avoid the revisitation of expressions while increasing the exploration capabilities.

\section{SymRegg: Equality Graph Assisted Symbolic Regression}\label{sec:symregg}

SymRegg\footnote{The name is the combination of Symbolic Regression and \emph{egg}, inspired on the e-graph library of the same name.} departs from the principle of being a simple search heuristic that efficiently explores the search space of symbolic expressions. Simplicity here means that it should perform as well as the current state-of-the-art while keeping the number of hyperparameter settings to a minimum.

The main idea of the algorithm is to incrementally insert expressions in the e-graph by exploiting the visited expressions so far. In the context of the algorithm, an expression is visited if it was already evaluated. An expression that is already contained in the e-graph but was never evaluated, is still considered as not visited. For example, upon inserting and evaluating the expression $x * \theta^x$, the sub-expressions $x$ and $\theta^x$ will not be evaluated and thus will remain labeled as unvisited. 

The algorithm starts by creating an empty e-graph that keeps the history of visited expressions during the search. Next, it inserts and evaluates all terminal tokens (i.e., variables and numerical parameter placeholder) to ensure the e-graph contains all possible terminals, followed by the insertion and evaluation of a random expression. Finally, it repeats the following two steps until the algorithm evaluates a total of $N$ expressions: i) attempts to generate a never visited expression, ii) if it succeeds, runs one step of equality saturation to generate the equivalent expressions. 

The first step relies on a sequence of attempts using different methods to try and generate a new expression. Since there is no guarantee of creating an unvisited expression, the expression is queried in the e-graph and, if it has never been visited, it will be evaluated, inserted into the e-graph and followed by one step of equality saturation. Applying equality saturation will generate some of the equivalent expressions of the newly inserted expression and their child expressions.
To create a new expression, the algorithm tries four different methods in the following order, stopping as soon as an unvisited expression is created:

\begin{enumerate}
    \item Create a new expression from one (perturbation) or two (recombination) expressions sampled from the set of the union of the $50$ most accurate expressions for each size $s$.
    \item Same as above but sampled from the set of the $100$ most accurate expressions.
    \item Evaluate a random not yet evaluated e-class.
    \item Insert a random expression.
\end{enumerate}

The first step start by selecting a set of the top $50$ expressions for every size $s$ among the visited expressions. For example, if the algorithm generated expressions with sizes $[1, 3, 5, 10]$, it will try to retrieve the top $50$ for each one of these sizes. From this set, it will apply a perturbation or a recombination with a probability of $50\%$ each. Depending on the choice of operation it will draw one or two expressions completely at random from this selection.
This first selection stimulates the combination of good expressions of different complexity, trying to improve the current Pareto front.
The second step is similar to the first  one but it simply samples the expressions from the top $100$ best expressions by accuracy. This incentivize the combination of the very best expressions.
The third step evaluates a not yet evaluated subtree by sampling a random e-class without evaluation information. Finally, the fourth step inserts a random expression being only executed in the event that every e-class in the e-graph is already evaluated.

The values of the set of the top expressions ($50$ and $100$, respectively) were empirically set to a value large enough to promote the creation of unvisited expressions, but not too large as to behave as a random search. Since the algorithm starts with a single expression, in the first step it will behave like a local search around the neighborhood of that solution and, as new expressions are introduced, it will converge to an equilibrium between exploration and exploitation.

\begin{figure}
    \centering
    \includegraphics[width=\linewidth]{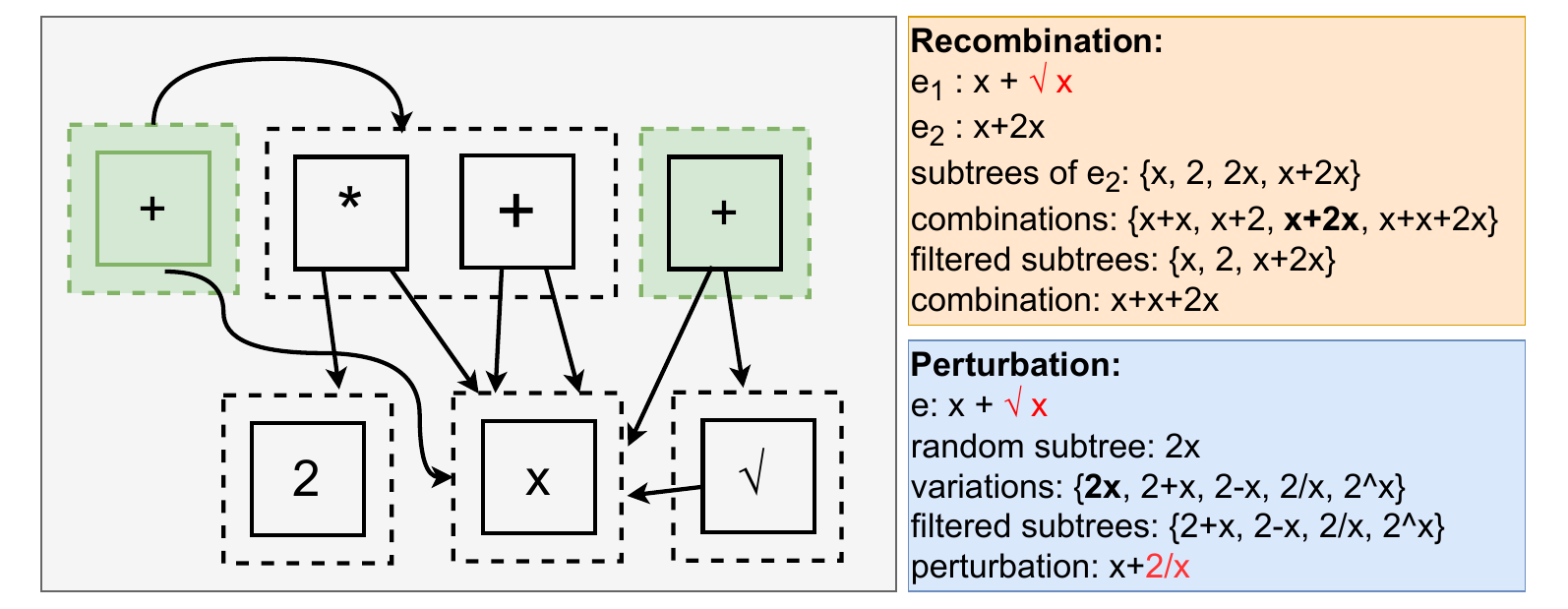}
    \caption{Example of the recombination and perturbation operators. In this e-graph the green e-classes represents the root of already evaluated expressions. This example will assume that the expression $x+\textcolor{red}{\mathbf{\sqrt{x}}}$ is sampled and the red highlight is the part of the expression to be replaced The recombination operator will sample the second expression, $x+2x$, that contains the set of subtrees $\{x+2x,x,2x,2\}$. From this set, the subtree $2x$ would create an already visited expression and then it is discarded before selecting the replacement. For the perturbation, supposing that replacing that part with a random subtree creates the expression $x+\textcolor{red}{\mathbf{2x}}$, which already exists in the e-graph, the algorithm will replace the multiplication operator with any other operator with the same arity that would generate an unvisited expression, such as $+,-,\div{},\textasciicircum$.
    }
    \label{fig:operators}
\end{figure}

The perturbation and recombination operators are the same as proposed in~\cite{eggp} (illustrated in Fig.~\ref{fig:operators}) exploiting the information stored on the e-graph to increase the probability of generating an unvisited expression. 
Essentially, these operators limit the set of choices for the recombination points or perturbation of a tree to enforce the creation of unvisited expressions. If these subsets are empty, it will return a failed attempt and continue to the next step of the algorithm.

Before evaluating the expression, the single step of equality saturation will insert into the e-graph the equivalent expressions and choose the smallest form to represent that set of equivalent expressions. So, for example, if the expression $\log{(\theta)} + \log{(x)}$ is inserted, after applying equality saturation, the e-graph will generate the expression $\log{(\theta x)}$ and use the latter to evaluate.

The set of terminals contain, besides the input variables, a single placeholder for the numerical parameters $\theta$. Whenever an expression is created with multiple parameter symbols, the multiple occurrences are relabeled sequentially to represent different parameters. For example, the expression $\theta x_0 + \log{(\theta - x_1)}$ would become $\theta_0 x_0 + \log{(\theta_1 - x_1)}$. The values of each $\theta_i$ are fitted to the data by optimizing one of the supported loss functions and stored into the e-graph. This fitting procedure is performed on a subset of the training data and the fitted expression is evaluated on the remaining data, called validation set.

During the equality saturation process, two or more already evaluated expressions can be found to be equivalent, in which case they are merged together. The algorithm will keep the fitness and parameter values information of the smallest of them.
This will ensure that the e-graph will always keep the information about the most compact representation of each evaluated expression.

Regarding the user-defined hyperparameters, the user can choose the total number of evaluated expressions, the maximum size of the expression, the ratio of the amount of data to use as training (the remainder is the validation), the loss function, the number of nonlinear optimization iterations for fitting, the number of retries for each fitting procedure, and the set of non-terminal symbols. It is also possible to set the maximum number $m$ of numerical parameters in the expression, in which case the set of terminals will contain the set $\{\theta_i \mid i = \{0 \ldots m\}\}$ allowing expressions to contain repeated occurring parameters, such as in $\theta_0 x_0 + \log{(\theta_1 x_1 + \theta_0)}$. Finally, it is possible to specify multiple training datasets with the same format representing samples from different populations, and the algorithm will search for a common parametric function that fits them all (similar to the concept of factor variables in regression~\cite{kronberger2018predicting}, but implemented as a multi-view symbolic regression~\cite{russeil2024multiview}). Different from some GP implementations, the influence of each one of these hyperparameters to the search are intuitive or they represent some desiderata about the searched equation. The number of evaluations and the maximum iterations of the parameter fitting procedure will increase the runtime but with a higher chance of finding a good quality expression. The ratio of training and validation data depends on the number of data points available, the maximum size of the expression specify the trade-off between accuracy and complexity as desired by the user, and the set of non-terminals should be set accordingly to some prior knowledge about the data. Comparing to GP, the user often have to decide the best equilibrium between number of iterations and population size to achieve the maximum number of evaluations without hurting the overall performance, another example is the unclear effect of the crossover and mutation rates, as they usually depends on the dataset and must be carefully set. Also different from GP, this algorithm does not rely on the selective pressure of methods such as tournament selection and it will naturally alternate between exploration and exploitation depending on the state of the e-graph, whether there is still many opportunities to create new expressions from the perturbations, or if it needs to insert random expressions to promote novelty. Another distinct characteristic is that SymRegg will only converge once it enumerates the entire search space, as it will keep trying to produce new solutions while rejecting redundant expressions. So, in the limit, it will behave as an inefficient enumeration algorithm, since it will \emph{generated} repeated solutions without evaluating them. 


\noindent\textcolor{red}{\textbf{Current limitations. }The main limitation of this approach is the possible overhead of maintaining the e-graph structure alongside the search but, as noted by de Fran\c{c}a \& Kronberger~\cite{eggp}, with a careful implementation this overhead is minimal. Additionally, the e-graph can grow exponentially with every iteration of equality saturation, this is alleviated here by running a single step for each new expression, which limits its growth while still being capable of detecting a subset of the equivalent expressions. This is also another limitation of \alg{}, as it will not be able to detect every equivalent expression since it will not run until saturation. Nevertheless, this will mostly limit overly complex expressions that would not be generated anyway due to expression size constraints (e.g., $x + x - x + x - x + \ldots$). Finally, the memory usage can grow to such an extent that it will reach a hardware limit, this can be solved in two ways: i) storing the e-graph into a database system where part of the information is on disk rather than in memory, ii) keeping only a subgraph of the top-$N$ expressions of the e-graph, allowing the algorithm to revisit some expressions.}

\noindent\textcolor{red}{\textbf{Related literature.} In a previous work~\cite{de2023reducing}, e-graphs, equality saturation, and SR were used together to verify whether popular SR implementations produced expressions with redundant parameters. The authors verified that, in different extent, this is a problem shared by the tested implementations. They also verified that simplifying the expressions with SymPy sometimes even increased the number of parameters. On the other hand, equality saturation was capable of simplifying the expressions removing almost every redundant parameter. As a followup work~\cite{kronberger2024jsc}, the authors verified that these redundant parameters could impact the convergence of the optimization method and hinder the search for an accurate expression.
The integration of e-graphs into the SR search was first proposed in~\cite{eggp} with the algorithm called \emph{eggp}, where the crossover and mutation operators were replaced by version that enforced the creation of unvisited expressions, as in this work. Different from \alg{}, the algorithm still relied on the selective pressure of applying tournament selection extracted from the current Pareto front. Unlike \alg{}, \emph{eggp} is subject to premature convergence as with most SR algorithms.
Finally, in~\cite{reggression}, the authors proposed a tool capable of exploring a database of symbolic models by allowing the user to retrieve top expressions filtered by patterns or other desiderata. }

\section{Experiments}\label{sec:experiments}

To evaluate the efficiency of SymRegg, we will follow the same experimental methods described in~\cite{kronberger2024inefficiency}. 
\textcolor{red}{The main objective is to measure with how many function evaluations each SR method achieves a prediction error below a specified threshold, measured as a percentage of the global optimum, for each dataset. As such, we measure the probability of achieving these thresholds within $1$ to $50\,000$ evaluations.} In this context, a single evaluation means generating an expression and fitting the parameters independent on the number of iterations necessary for the optimization. The probability will be calculated over $30$ runs of each algorithm on four different real-world datasets using two different maximum size ($10$ and $12$). Each algorithm was modified to print every visited expression in the same order they were generated. The probability values will be plotted considering different reasonable thresholds for the error measurements. These plots show the speed with which each algorithm reaches a high probability of succeeding in the search.

Besides SymRegg, we will test Tiny-GP\cite{Sipper2019tinyGP}, a minimalist implementation of genetic programming for symbolic regression adapted to perform parameter fitting, Operon~\cite{burlacu2020operon}, a high-performant implementation with competitive accuracy with opaque regression models, and a random sample of the enumerated space of expressions up to the maximum allowed size without replacement as in~\cite{kronberger2024inefficiency}. This last algorithm is equivalent to a permutation of the enumeration of the search space, stopping at any time
(PAES -- Permuted Anytime Exhaustive Search).

Our implementation of PAES starts with the enumeration of the search space following an improved version of Exhaustive Symbolic Regression (ESR)~\cite{bartlett2023exhaustive} as described in~\cite{kronberger2024inefficiency}. With the enumerated set of expressions, we sample expressions at random sequentially removing them from the set once they are drawn. This method simulates an ideal random search that cannot visit the same solution more than once, achieving a maximum efficiency in the exploration of small search spaces, as noted by Kronberger et al.~\cite{kronberger2024inefficiency}. \textcolor{red}{PAES is both optimal and complete, because it is guaranteed to find the global optima with an unlimited amount of time. On the other hand, local search is neither, but it usually performs better than PAES when the resources are limited~\cite{norvig2002modern}. In this setting, an algorithm that underperforms against PAES is not capable of exploiting the locality information essential to converge to a local optima.  We should stress that the maximum sizes used in these experiments are because of the limitation of the ESR. SymRegg, Tiny-GP, and Operon are not limited by these constraints.}
The datasets used for these experiments are:

\begin{itemize}
    \item \textbf{Beer's law:} attenuation of light to the properties of the material through which the light is traveling, known as Beer's law, extracted from~\cite{russeil2024multiview}.
    \item \textbf{Nikuradse 1:} this dataset describes the flow in rough pipes using the ratio of the relative roughness and the logarithm of the Reynolds number~\cite{nikuradse1950,Guimera2020}. 
    \item \textbf{Nikuradse 2:} same as above, but using a single input and a transformed target variable.
    \item \textbf{Supernovae:} brightness of light curves of a supernova event, extracted from~\cite{russeil2024multiview}.
\end{itemize}

The hyperparameters values for Operon and Tiny-GP were empirically chosen based on previous experiments with these datasets (see Table~\ref{tab:hp})\cite{kronberger2024inefficiency}. We restricted the set of non-terminals, total number of evaluations, and loss function to be the same. For SymRegg, we set the optimization iterations and retries to $100$ and $2$, respectively. For PAES, we set those values to $1\,000$ and $50$, which gives a higher probability of obtaining the global minimum during the fitness process. Even though this gives an advantage to PAES, these values were chosen so that PAES is closer to an upper limit of the performance. \textcolor{red}{We should also notice that Operon automatically applies scaling to the leaf nodes without counting them towards the size limit, thus sometimes generating longer expressions than the specified maximum size. For the experiments, we have removed all expressions that exceeds the limits before generating the plots and tables.}

\begin{table}[!t]
\caption{Choice of hyperparameters for each algorithm, a dash indicates that the parameter is not applicable to that algorithm. Notice that the \emph{optimization iterations} hyperparameter refers to the maximum number of iterations, so the methods may stop earlier on convergence.}
\label{tab:hp}
\begin{tabular}{lllll}
\toprule
Hyperparameter & Operon & Tiny-GP & SymRegg & PAES \\
\midrule
population size & $1\,000$ & $1\,000$ & -- & --\\
generations & $50$ & $50$ & -- & --\\
min. depth &  $1$ & $2$ & -- & --\\
max. depth & $10$ & $4$ & -- & --\\
tournament size & $2$ & $5$ & -- & --\\
prob. crossover & $1.0$ & $1.0$ & -- & --\\ 
prob. mutation & $0.15$ & $0.25$ & -- & --\\
linear scaling & -- & False & -- & --\\
optimization iterations & $100$ & $100$ & $2 \times 100$ & $50 \times 1\,000$\\
optimizer & LM~\cite{levenberg1944method} & L-BFGS\cite{liu1989limited} & Var1~\cite{vlcek2006shifted} & Var1~\cite{vlcek2006shifted} \\
\toprule 
Hyperparameter & all algorithms &  &  &  \\
\midrule 
non-terminal set & $+,-,*,/,\operatorname{powabs},\operatorname{recip}$ &  &  & \\
total evaluations & $50\,000$ &  &  & \\
loss function & MSE &  &  & \\
folds & 1 & & & \\ 
\# of free parameters & unlimited & & & \\ 
\botrule
\end{tabular}
\vspace*{-4pt}
\end{table}

\section{Results}\label{sec:results}

Before exploring the obtained results, Table~\ref{tab:esr-qualities} show some statistics of the enumeration of the search space for each dataset. In this table we can see different reference thresholds for the MSE and the percentage of expressions with that value or better. From this table we can understand the difficulty of each dataset. For Beer's law and Supernovae, only $0.01\%$ of the expressions of size $10$ are of the same order of magnitude as the best expression. For Nikuradse 2, $0.1\%$ of the expressions are at the same order as the best one, while Nikuradse 1 seems to be more challenging with only $0.001\%$ of the expressions of the same order as the best.

\begin{table}[!th]
\centering
\caption{Errors of best solutions found with ESR. Freq. is the ratio of all expressions (for maximum length 10 and 12) with an MSE value better of equal to the value reported in the last column.}\label{tab:esr-qualities}
\begin{tabular}{lccc|ccc}
\toprule 
  Dataset & Len. & Freq. & MSE & Len. & Freq. & MSE \\ \midrule
    Beer        & 10 & Best    & $2.463 \cdot 10^{-5}$ & 12 & Best    & $1.402 \cdot 10^{-5}$ \\  
                &    & 0.01\%  & $5.652 \cdot 10^{-5}$ &    & 0.01\%  & $2.394 \cdot 10^{-5}$\\  
                &    & 0.1\%   & $1.866 \cdot 10^{-4}$ &    & 0.1\%   & $6.44 \cdot 10^{-5}$ \\  
                &    & 1\%     & $1.505 \cdot 10^{-3}$ &    & 1\%     & $7.354 \cdot 10^{-4}$ \\ \midrule
    Supernovae  & 10 & Best    & $5.647 \cdot 10^{-4}$ & 12 & Best    & $4.108 \cdot 10^{-4}$\\  
                &    & 0.01\%  & $7.155 \cdot 10^{-4}$ &    & 0.01\%  & $6.261 \cdot 10^{-4}$\\  
                &    & 0.1\%   & $1.837 \cdot 10^{-3}$ &    & 0.1\%   & $1.875 \cdot 10^{-3}$\\  
                &    & 1\%     & $1.663 \cdot 10^{-2}$ &    & 1\%     & $1.646 \cdot 10^{-2}$ \\ \midrule 
    Nikuradse 1 & 10 & Best    & $9.339 \cdot 10^{-4}$ & 12 & Best    & $5.057 \cdot 10^{-4}$\\  
                &    & 0.01\%  & $1.090 \cdot 10^{-3}$ &    & 0.01\% & $9.825 \cdot 10^{-4}$\\  
                &    & 0.1\%   & $1.296 \cdot 10^{-3}$ &    & 0.1\%  & $1.239 \cdot 10^{-3}$\\  
                &    & 1\%     & $1.570 \cdot 10^{-3}$ &    & 1\%   & $1.656 \cdot 10^{-3}$\\ \midrule 
                
    Nikuradse 2 & 10 & Best    & $2.463 \cdot 10^{-3}$ & 12 & Best    & $1.367 \cdot 10^{-3}$\\  
                &    & 0.01\%  & $4.812 \cdot 10^{-3}$ &    & 0.01\%  & $2.174 \cdot 10^{-3}$\\  
                &    & 0.1\%   & $8.259 \cdot 10^{-3}$ &    & 0.1\%   & $4.538 \cdot 10^{-3}$\\  
                &    & 1\%     & $1.736 \cdot 10^{-2}$ &    & 1\%     & $1.12 \cdot 10^{-2}$\\ \midrule 
                \botrule 
\end{tabular}
\end{table}

\begin{table}
\caption{Mean and standard deviation of MSE obtained in $30$ runs of every algorithm.}
\label{tab:mse}
\centering
\begin{tabular}{l|c|c|c}
\toprule
 Dataset & SymRegg & Tiny-GP & Operon \\
\midrule
Beer (10) & $\mathbf{1.25 \cdot 10^{-4}} \pm 4.35 \cdot 10^{-5}$ & $1.27\cdot 10^{-5}$ $\pm 2.96\cdot 10^{-5}$ & $2.76\cdot 10^{-4}$ $\pm 1.22\cdot 10^{-4}$ \\
Beer (12) & $6.40\cdot 10^{-5}$ $\pm 1.62\cdot 10^{-5}$ & $5.05\cdot 10^{-5}$ $\pm 3.96\cdot 10^{-5}$ & $\mathbf{5.18\cdot 10^{-5}} \pm 2.25\cdot 10^{-5}$ \\
Niku. 1 (10) & $1.03\cdot 10^{-3}$ $\pm 6.61\cdot 10^{-5}$ & $1.30\cdot 10^{-3}$ $\pm 4.09\cdot 10^{-6}$ & $\mathbf{9.86\cdot 10^{-4}} \pm 1.41\cdot 10^{-7}$ \\
Niku. 1 (12) & $\mathbf{9.11\cdot 10^{-4}} \pm 5.97\cdot 10^{-5}$ & $1.29\cdot 10^{-3}$ $\pm 8.30\cdot 10^{-6}$ & $9.38\cdot 10^{-4}$ $\pm 2.29\cdot 10^{-5}$ \\
Niku. 2 (10) & $\mathbf{4.06\cdot 10^{-3}} \pm 3.74\cdot 10^{-4}$ & $5.17\cdot 10^{-3}$ $\pm 9.67\cdot 10^{-4}$ & $1.91\cdot 10^{-2}$ $\pm 4.22\cdot 10^{-3}$ \\
Niku. 2 (12) & $\mathbf{1.72\cdot 10^{-3}} \pm 9.41\cdot 10^{-5}$ & $3.47\cdot 10^{-3}$ $\pm 1.01\cdot 10^{-3}$ & $5.06\cdot 10^{-3}$ $\pm 1.39\cdot 10^{-3}$ \\
Super. (10) & $\mathbf{7.15\cdot 10^{-4}} \pm 3.05 \cdot 10^{-5}$ & $2.30\cdot 10^{-3}$ $\pm 2.75\cdot 10^{-3}$ & $2.21\cdot 10^{-3}$ $\pm 1.16\cdot 10^{-3}$ \\
Super. (12) & $\mathbf{5.70\cdot 10^{-4}} \pm 5.14 \cdot 10^{-5}$ & $9.81\cdot 10^{-4}$ $\pm 6.89\cdot 10^{-4}$ & $6.64\cdot 10^{-4}$ $\pm 7.17\cdot 10^{-5}$ \\
\botrule 
\end{tabular}
\end{table}

\begin{figure}[!th]
\centering
\includegraphics[width=1in,trim={1.3cm 4.6cm 3.5cm 0},clip]{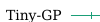}
\includegraphics[width=1in,trim={1.3cm 4.3cm 3.4cm 0.4cm},clip]{figs/beer_len10.pdf}
\includegraphics[width=1in,trim={1.2cm 4.0cm 3.4cm 0.7cm},clip]{figs/beer_len10.pdf}
\includegraphics[width=0.8in,trim={1.6cm 3.7cm 3.3cm 1.0cm},clip]{figs/beer_len10.pdf} \\
\subfloat[Beer's law - size 10]{\includegraphics[width=2.8in,trim={0 0 0cm 1.4cm},clip]{figs/beer_len10.pdf}}
\subfloat[Beer's law - size 12]{\includegraphics[width=2.6in,trim={0.5cm 0 0 1.4cm},clip]{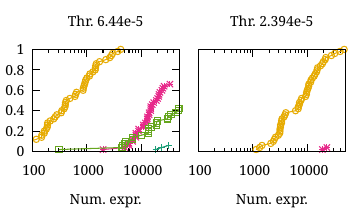}} \\
\subfloat[Supernovae - size 10]{\includegraphics[width=2.8in,trim={0 0 0cm 1.4cm},clip]{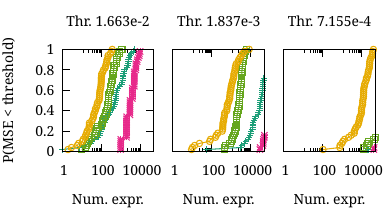}}
\subfloat[Supernovae - size 12]{\includegraphics[width=2.6in,trim={0.5cm 0 0 1.4cm},clip]{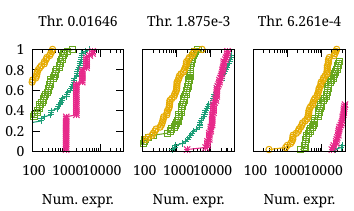}} \\
\subfloat[Nikuradse 2 - size 10]{\includegraphics[width=2.8in,trim={0 0 0cm 1.4cm},clip]{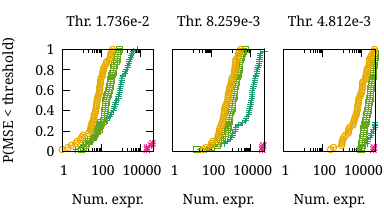}}
\subfloat[Nikuradse 2 - size 12]{\includegraphics[width=2.6in,trim={0.5cm 0 0 1.4cm},clip]{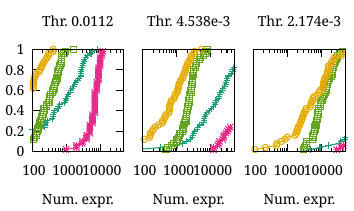}} \\
\subfloat[Nikuradse 1 - size 10]{\includegraphics[width=2.8in,trim={0 0 0cm 1.4cm},clip]{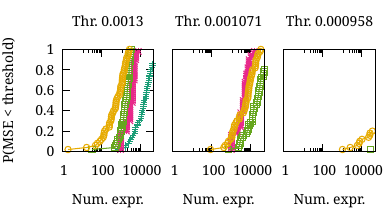}}
\subfloat[Nikuradse 1 - size 12]{\includegraphics[width=2.6in,trim={0.5cm 0 0 1.4cm},clip]{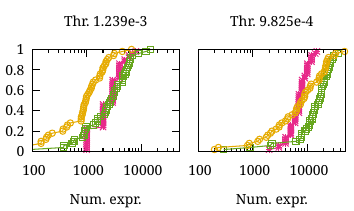}}
\caption{Probability of achieving a solution with MSE value below a given threshold (top) after a number of evaluations.}
\label{fig:results}
\end{figure}

In Figure~\ref{fig:results}, we can see the plots of the probability of achieving the specified threshold (displayed at the top of each plot) after a certain number of visited expressions. From these figures we can see that, except for the Beer's law dataset, SymRegg is always close to the performance plot of PAES, showing that the efficiency is close to an idealized algorithm. This becomes even more evident when we compare it against Tiny-GP, which often requires $10$ times more evaluations than SymRegg to achieve the same performance. In some cases Tiny-GP cannot even achieve the specified threshold. We should notice that this version of Tiny-GP is equivalent to SymRegg without the use of e-graphs. 
Finally, when compared to Operon, we can see that SymRegg is only worse at some settings of Nikuradse 1, specifically on smaller thresholds. In some situations, Operon either failed to achieve the specified threshold or it did take longer than SymREgg.

Table~\ref{tab:mse} shows the average MSE obtained by the best expression of each experiment. We can see from this table that, on average, SymRegg obtained the best solutions given the expression size limits. In Beer's law (12) and Nikuradse 1 (10), its average is close to the best (Operon). Table~\ref{tab:runtime} shows the average runtime for each algorithm in the different experiments. From this table, we can see that \alg{} stays in between Tiny-GP and Operon. Operon is a highly-tuned, multi-threaded implementation, while \alg{} is single-threaded with a comparable performance to other traditional GP implementations, Tiny-GP is implemented in pure Python which leads to a much slower performance than the other two algorithms. 
\textcolor{red}{Taking the best expression found by each algorithm on the Nikuradse 2 dataset in the first run, we have:}

\begin{align*}
  & f_{\text{SymRegg}} = |\theta_0|^x + |x|^{\frac{1}{x} + \theta_1} & f_{\text{Operon}} &= |\theta_0|^{\theta_1 x} (\theta_2 + \theta_3 x) + \theta_4 x \\
  & f_{\text{Tiny-GP}} = \theta_0 - \left|\frac{1}{x - \theta_1} + \theta_2\right|^{\theta_3} & f_{\text{PAES}} &= \frac{1}{\theta_0 - \left| \theta_1 x \right|^{\left| \theta_2 \right|^x}} 
\end{align*}
\textcolor{red}{since the experiments were performed with a limited maximum size, they are all close to the limit. The only difference in complexity is their functional form and the number of free parameters. Operon and Tiny-GP seems to prefer models with additional degree-of-freedoms (i.e., more parameters) tend to create an excess of parameters in their expressions, more than needed when compared to the optimal expression (PAES). SymRegg on the other hand is more conservative in the placement of parameters.}

SymRegg is available at \url{https://github.com/folivetti/symregg} and it is also installable via \emph{pip}. The result files for this experiment and additional information can be found at \url{https://github.com/folivetti/symregg_royalsociety}.

\begin{table}[!t]
    \centering
    \caption{Average runtime (in secs.) for each dataset (size in parenthesis) with each tested maximum size. This version of Tiny-GP is implemented in Python, explaining the larger runtimes.}
    \label{tab:runtime}
    \begin{tabular}{c|c|c|c|c|c|c|c}
    \toprule
        Dataset & \alg{} & Operon & Tiny-GP & Dataset & \alg{} & Operon & Tiny-GP \\ \midrule 
        Beer (10) & 63.87 & 1.84 & 265.87 & Niku 2 (10) & 115.33 & 2.59 & 221.03\\
        Beer (12) & 52.93 & 2.36 & 287.63 & Niku 2 (12) & 119.8 & 3.13 & 263.87\\ 
        Super. (10) & 69.77 & 1.36 & 203.53 & Niku 1 (10) & 76.73 & 2.38 & 324.53\\ 
        Super. (12) & 62.07 & 1.66 & 228.03 & Niku 1 (12) & 80.93 & 3.17 & 369.73\\ 
         
        \botrule 
    \end{tabular}
    
\end{table}

\vspace*{-5pt}

\section{Conclusion}\label{sec:conclusion}

Equation discovery has become an important method that helps fundamental sciences to understand new phenomena. This task is particularly challenging due to the lack of smoothness of the search space and abundance of redundant solutions. This is in part due to the inability to verify if a given expression was already visited during the search in one of their many equivalent forms. 
In this paper, we advance in this challenge with the use of equality graphs to efficiently store the visited expressions while exploiting its innate query system to enforce, when possible, the creation of unvisited expressions during the search. This algorithm, called \alg{}, sequentially tries to perform one of four moves: i) perturb solutions from a sample of a mix of the top expressions of different lengths; ii) perturb solutions from a sample of the top expressions; iii) evaluate unevaluated sub-expressions; iv) insert a random expression. These steps are tried in order until an unvisited expression is generated, this creates a natural shift from exploitation (perturbation and recombination), local search (evaluation of sub-expressions), and exploration (insertion of random expressions) that depends solely on the current state of the e-graph, or how easily it is to generate a novel expression.
Different from the traditional genetic programming algorithm, it does not rely on the concept of a \emph{population}, but it uses a subset of the visited expressions as a region of interest. It also does not rely on a strict selective pressure, such as tournament selection, using the current state of the e-graph to control the balance between exploration-exploitation.
\textcolor{red}{\alg{} is closely related to an idealized search (i.e., search algorithm that always sample a new expression) as it can offer a limited guarantee of generating unvisited expressions through its  perturbation procedure or the evaluation of subtrees. This differs from PAES as the new expressions are created from previously visited solutions (exploitation) and, in certain situations, it will not be able to produce new expressions. As such, it is expected that \alg{} performs closely to PAES but, the fact that PAES still achieved slightly better results means that the perturbation methods still fail to efficiently exploit the locality of the search space.}

We performed experiments comparing \alg{} with a high-performant symbolic regression algorithm (Operon), a minimalist implementation of GP (Tiny-GP), and an ideal random search.  The results indicated that the use of e-graph helped to achieve a performance close to the ideal upper bound defined by the random search.

For future steps, we will investigate whether the size of the selection of the top expressions have any influence in the overall performance of the algorithm and if a self-adaptive mechanism would improve the exploration-exploitation balance. Also, as the e-graph implements a pattern matching system, it is possible to extract a general \emph{template} of the data generating function that can later be related to common first-principle models.


\ack{F.O.F. is supported by Conselho Nacional de Desenvolvimento Cient\'{i}fico e Tecnol\'{o}gico (CNPq) grant 301596/2022-0.
G.K. is supported by the Austrian Federal Ministry for Climate Action, Environment, Energy, Mobility, Innovation and Technology, the Federal Ministry for Labour and Economy, and the regional government of Upper Austria within the COMET project ProMetHeus (904919) supported by the Austrian Research Promotion Agency (FFG). }

\bibliographystyle{plain}
\bibliography{references}

\end{document}